\documentclass[a4paper,fleqn]{cas-dc}

\usepackage[authoryear]{natbib}
\usepackage{soul} %
\usepackage{color, xcolor} %
\usepackage{pifont}
\usepackage[capitalize]{cleveref}
\usepackage{subfigure}
\usepackage[normalem]{ulem}
\useunder{\uline}{\ul}{}
\usepackage{caption}
\usepackage{graphicx}
\usepackage{booktabs}
\def\tsc#1{\csdef{#1}{\textsc{\lowercase{#1}}\xspace}}
\tsc{WGM}
\tsc{QE}

\usepackage{amsmath,amsfonts}
\usepackage{algorithm}
\usepackage{algorithmic}
\usepackage{array}
\usepackage{textcomp}
\usepackage{stfloats}
\usepackage{url}
\usepackage{verbatim}
\usepackage[table]{xcolor}
\usepackage{graphicx}
\usepackage{booktabs}
\usepackage{multirow}
\usepackage{bbding}
\usepackage{caption}
\usepackage{subcaption}
\usepackage{tabularx, booktabs}

\newcommand{\etal}{\textit{et}~\textit{al}.} 
\usepackage{paralist}
\usepackage{tabularx}

\begin{document}
\let\WriteBookmarks\relax
\def\floatpagepagefraction{1}
\def\textpagefraction{.001}

\shorttitle{Boosting Point-supervised Temporal Action Localization via Text Refinement and Alignment}

\shortauthors{Yunchuan Ma et al.}  

\title [mode = title]{Boosting Point-supervised Temporal Action Localization via Text Refinement and Alignment}

\author[1]{Yunchuan Ma}
[orcid=0009-0009-5042-4198]
\ead{mayunchuan23@mails.ucas.ac.cn}

\author[1]{Laiyun Qing}
[orcid=0000-0001-9923-5034]
\cormark[1]
\ead{lyqing@ucas.ac.cn}

\author[1]{Guorong Li}
[orcid=0000-0003-3954-2387]
\ead{liguorong@ucas.ac.cn}

\author[1]{Yuqing Liu}
[orcid=0009-0004-1883-051X]
\ead{liuyuqing231@mails.ucas.ac.cn}

\author[2]{Yuankai Qi}
[orcid=0000-0003-4312-5682]
\ead{yuankai.qi@mq.edu.au}

\author[1]{Qingming Huang}
[orcid=0000-0001-7542-296X]
\ead{qmhuang@ucas.ac.cn}

\cortext[cor1]{Corresponding author}

\address[1]{University of Chinese Academy of Sciences, Beijing,100190, China}
\address[2]{Macquarie University}

\credit{}

\begin{abstract}
Recently, point-supervised temporal action localization has gained significant attention for its effective balance between labeling costs and localization accuracy.
However, current methods \textcolor{black}{primarily rely on visual features and do not fully exploit the complementary semantic information contained in textual descriptions}.
To address this issue, we propose a Text Refinement and Alignment (TRA) framework that \textcolor{black}{incorporates refined textual semantics to complement visual representations for point-supervised temporal action localization}.
This is achieved by designing two new modules for the original point-supervised framework:
a Point-based Text Refinement module (PTR) and a Point-based Multimodal Alignment module (PMA).
Specifically, we first generate descriptions for video frames using a pre-trained multimodal model.
Next, PTR refines the initial descriptions by leveraging point annotations together with multiple \textcolor{black}{fixed} pre-trained models.
PMA then projects \textcolor{black}{the visual and textual features} into a unified semantic space and \textcolor{black}{employs a point-based multimodal contrastive learning objective} to reduce the gap between visual and linguistic modalities.
\textcolor{black}{Finally, the aligned multimodal features are fed into the action detector for temporal action localization.}
\textcolor{black}{Extensive experiments on five widely used benchmarks demonstrate that TRA consistently improves strong point-supervised baselines and achieves competitive performance compared with state-of-the-art methods.}
\textcolor{black}{In particular, on THUMOS'14, TRA achieves 58.5\% AVG mAP@[0.1:0.7], outperforming the visual-only baseline by 2.3\%.
}
\textcolor{black}{These results demonstrate that refined textual semantics can effectively complement visual representations and improve temporal action localization performance.}
\end{abstract}


\maketitle



\section{Introduction}
\label{sec:intro}
{T}{emporal} Action Localization aims to localize and classify action instances in a long untrimmed video, which plays an essential role in various applications~\citep{eswa4,eswa5}, such as highlight detection, security monitoring, and action spotting.
Traditional fully-supervised methods~\citep{eswa2} have achieved high performance but require high-quality temporal boundary annotations, which are time-consuming and rarely readily available.
In contrast, weakly-supervised temporal action localization (WTAL)~\citep{eswa1}, only requires video-level annotations~(\textit{i.e.}, only label which action categories are contained in a video). 
However, the lack of instance-level annotations in WTAL makes it challenging for models to distinguish actions from backgrounds, leading to inferior performance compared to fully-supervised methods.

To balance annotation labor cost and detection performance, point-supervised temporal action localization (PTAL)~\citep{eswa3} is proposed, requiring only one single frame label for each action instance.
As the pioneer, 
SF-Net~\citep{SF-Net} 
iteratively mines pseudo action frames and background frames to provide supplementary supervision during training.
Following this setting, many researchers further propose various optimization methods.
For example, LACP~\citep{LACP} designs a greedy algorithm to generate the dense optimal sequence and learns the completeness of the action instances from point annotations.
TSPNet~\citep{TSPNet} leverages the prior knowledge that humans tend to annotate at the center of instances, and learns a center score to mitigate the misalignment issue.
To alleviate the semantic ambiguity in determining action boundaries, SMBD~\citep{SMBD} proposes a stepwise multi-grained boundary detector to locate the action boundaries.
Meanwhile, HR-Pro~\citep{HR-Pro} proposes a two-stage learning framework, which learns the discriminative snippet-level scores to generate more reliable proposals and adjusts these proposals through instance-level feature learning.
\textcolor{black}{Recently, QROT introduces query reformation and optimal transport into point-supervised TAL to improve proposal generation and matching.}

\begin{figure}
\centering
\includegraphics[width=\linewidth]{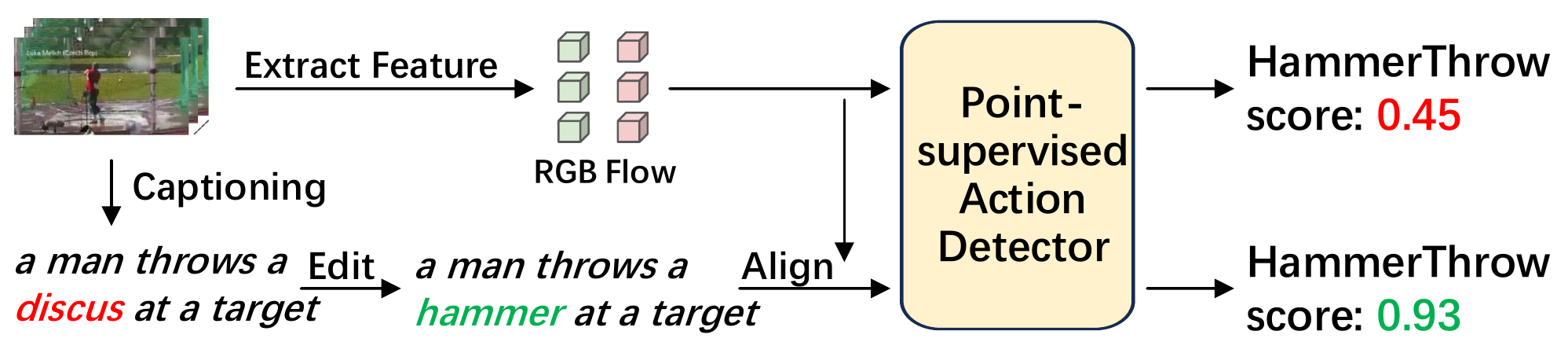}
\caption{Previous approaches (upper branch) directly use visual features as the input to the detector, whereas our method (lower branch) first generates visual descriptions, then edits them for improved accuracy, and finally aligns multiple modalities as input to the detector.
}
\label{fig:insight}
\end{figure}

\textcolor{black}{
Although existing PTAL methods have made substantial progress, their utilization of textual semantic information remains limited. Recent methods, such as QROT~\citep{QROT}, have explored semantic query learning or label-derived semantic priors. However, these methods mainly rely on annotation-level action semantics provided by point labels or action category labels, while fine-grained textual-modality information from visual descriptions remains underexplored. To fully exploit high-level semantics, it is beneficial to incorporate relevant textual information. To this end, we propose a novel framework, termed Text Refinement and Alignment (TRA). Different from QROT, which mainly exploits label-derived semantic priors, TRA leverages refined textual descriptions generated from visual content and further aligns the relevant visual and linguistic modalities, thereby improving the precision of action localization.
}
%
As shown in the lower part of Figure~\ref{fig:insight}, unlike previous methods, we plug external visual descriptions as input and obtain more accurate class activation scores.
Specifically, we begin by following traditional methods to segment the videos into dense short snippets, and then utilize an off-the-shelf caption generator (\textit{i.e.}, BLIP2) to produce detailed captions for these snippets.
Although powerful pre-trained captioning models can generate fluent captions, 
they might generate incorrect captions for different but similar actions (\textit{e.g.}, throw a discus and throw a hammer).
To alleviate the issue, we propose a Point-based Text Refinement (PTR) module to edit and enhance these sentences. Specific details of this module are provided later in Sec.~\ref{ptr}.
Subsequently, we extract features from different modalities and feed them into our proposed Point-based Multimodal Alignment (PMA) module for aligning. In PMA, we design a point-based multimodal contrastive learning loss to minimize the gap between different modalities of the same action as much as possible and increase the disparity between different action classes.

In summary, the main contributions of this work are as follows:
\begin{compactitem}
\item \textcolor{black}{We identify the neglect of textual semantics in point-supervised temporal action localization and address it by leveraging point annotations to mine accurate textual semantics and fully exploit their potential.}
\item We propose a Text Refinement and Alignment (TRA) framework, which effectively refines general captions and aligns vision and language modalities, enhancing semantic information to improve localization precision.
\item We design two new modules: the Point-based Text Refinement (PTR) module and the Point-based Multimodal Alignment (PMA) module. The former edits incorrect captions into correct ones with point annotations. The latter aligns visual and text modalities using predicted pseudo-point labels.
\item 
Extensive experiments on the THUMOS'14, GTEA, BEOID, ActivityNet1.2, and ActivityNet1.3 datasets
demonstrate the favorable performance of the proposed method compared to several state-of-the-art approaches.
Furthermore, the computational overhead of our framework is moderate, and it can be executed on a single consumer-grade GPU (\textit{e.g.}, NVIDIA GeForce RTX-3090 with 24 GB memory), 
\textcolor{black}{
demonstrating its potential for offline applications where textual features can be precomputed.
}
\end{compactitem}

\section{Related Work}
\label{sec:related_work}
\subsection{Temporal Action Localization}
\noindent\textbf{Fully-Supervised Temporal Action Localization} (FTAL), requires precise annotations of the start and end times of action instances in untrimmed videos.
Similar to the categorization of object detection methods~\citep{yolo,faster_rcnn}, mainstream FTAL methods can be divided into one-stage and two-stage approaches based on the localization workflow.
The one-stage methods~\citep{gtea,actionformer} utilize an end-to-end framework to locate and classify action instances in untrimmed videos.
In contrast, the two-stage methods~\citep{BSN,BMN}, first densely generate proposals, then predict the categories of these proposals and adjust their boundaries.
Although fully-supervised methods achieve impressive results, they rely on high-precision manual annotations, making them very costly.

\noindent\textbf{Weakly-Supervised Temporal Action Localization} (WTAL) only uses video-level labels to detect action instances.
Some Multiple Instance Learning (MIL) based methods ~\citep{P-MIL} generate class activation sequences and localize action instances via threshold-based processing.
Some other methods focus on feature learning to acquire more discriminative features.
For example, DCC~\citep{DCVC} designs a network that reduces background noise to better identify the action instances in a video.
Additionally, some efforts attempt to generate pseudo-labels to enhance supervision information.
TSCN~\citep{TSCN} creates frame-level pseudo-labels by blending attention sequences after considering consensus from two sources.
Although weakly-supervised methods significantly reduce annotation costs, they also hinder performance.

\noindent\textbf{Point-Supervised Temporal Action Localization} (PTAL) serves as a middle ground, bridging the gap between weak supervision and full supervision, while requiring only one-sixth the annotation time (\textcolor{black}{50s} \textit{vs.} 300s per 1-min video)~\citep{SF-Net, LACP} of fully supervised methods.
SF-Net~\citep{SF-Net} first introduces single-frame supervision to the temporal action localization task, iteratively mining pseudo action frames and background frames to provide more supervision during training.
Following this paradigm, numerous researchers study various optimization strategies. For instance, LACP~\citep{LACP} utilizes a greedy algorithm to generate densely optimized sequences, learning the completeness of action instances from point annotations.
TSPNet~\citep{TSPNet} takes advantage of the prior knowledge that annotations are typically made at the center of action instances, developing a center scoring mechanism to improve localization accuracy.
To address the issue of semantic ambiguity in determining action boundaries, SMBD~\citep{SMBD} develops a step-by-step multi-grained boundary detection method to precisely locate action boundaries.
Besides, HR-Pro~\citep{HR-Pro} first generates more reliable proposals by learning discriminative snippet-level scores and adjusts these proposals through instance-level feature learning.
\textcolor{black}{
However, most existing point-supervised temporal action localization methods mainly rely on visual features and sparse point annotations, while the fine-grained semantics contained in visual descriptions remain underexplored. Some recent methods, such as QROT~\citep{QROT}, introduce label-derived semantic priors, but they do not explicitly exploit textual descriptions generated from visual content. In contrast, TRA refines frame-level descriptions using point annotations and further aligns textual and visual representations, providing complementary semantic cues for more accurate localization.
}

\subsection{Zero-Shot Image Captioning}
Recently, zero-shot image captioning has gained more and more attention.
Some methods~\citep{ZeroCap, MAGIC} use a pre-trained vision-language model to guide the generation of a pre-trained LM.
Specifically, ZeroCap~\citep{ZeroCap} combines the CLIP~\citep{clip} and GPT-2~\citep{gpt2} to generate image captions through the gradient-search iteration.
On the other hand, Some methods~\citep{blip2,llava} fine-tune a few parameters on large-scale data to facilitate better integration of image encoders with large language models.
For example, BLIP-2~\citep{blip2} employs a lightweight perceiver-based transformer (\textit{i.e.}, Q-former) to bridge a frozen visual encoder and a large language model.
In this work, we choose the widely-used BLIP2 as our caption generator.

\begin{figure*}[H]
\centering
\includegraphics[width=\linewidth]{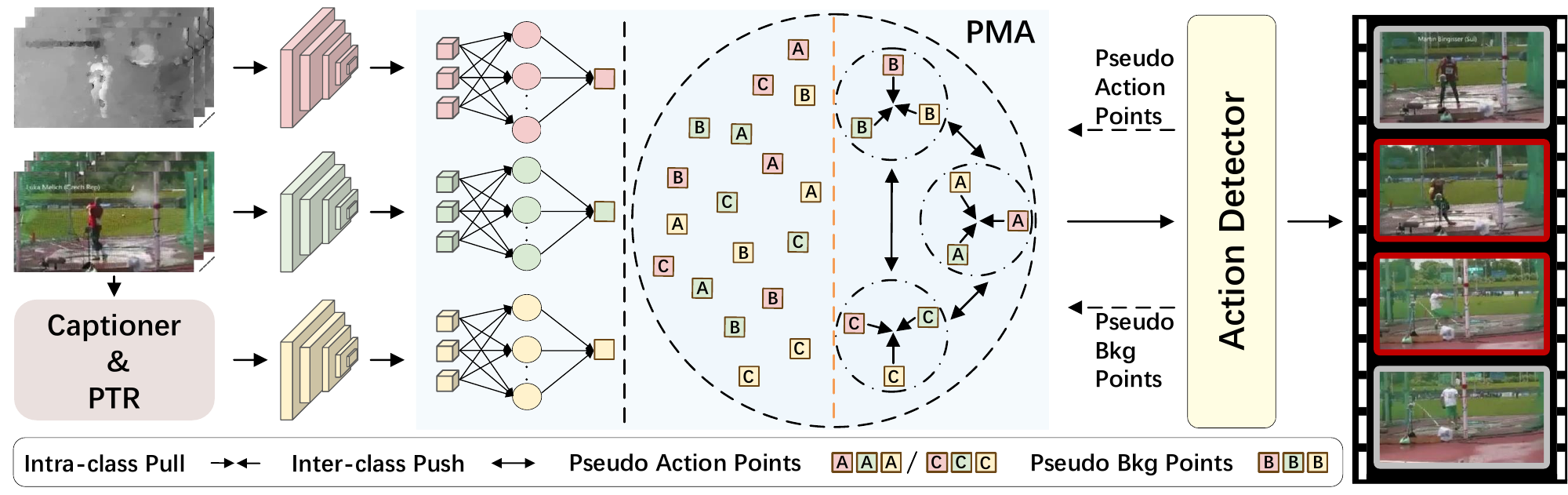}
\caption{Overview of the proposed TRA. 
Differing from conventional methods that depend solely on visual features (such as RGB and optical flow), our method first employs a pre-trained captioner to generate relevant descriptions, then refines these descriptions through PTR, and maps all features to the same dimension using a linear layer. Subsequently, the detector’s predicted pseudo action and background points are used to select the corresponding features for contrastive learning, achieving feature alignment. Ultimately, the aligned features are fed into the action detector for localization.
The detailed structure of PTR is provided in Sec.~\ref{ptr}.
}
\label{fig:method}
\end{figure*}

\section{Method}
In this section, we first describe the task definition and present the overall architecture of our method.
Then, we detail the point-based text refinement module (Sec.~\ref{ptr}) and point-based multimodal alignment module (Sec.~\ref{pma}), respectively.
In the end, we present the training and inference process in Sec.~\ref{training} and Sec.~\ref{inference}.

\vspace{1mm}
\noindent\textbf{Task Definition.}~\label{task definition}
Point-supervised Temporal Action Localization (PTAL) aims to learn a model
using merely a single-frame annotation per action instance. 
Each action instance is annotated with a timestamp $t$ and its category $y$. 
Given an untrimmed video, a trained PTAL model predicts actions in the form $(\hat{s}, \hat{e}, \hat{y}, con)$, where $\hat{s}$ and $\hat{e}$ denote the start and end times of each predicted action instance, $\hat{y}$ is the predicted action category, and $con$ is the confidence score.

\vspace{1mm}
\noindent\textbf{Method Overview.}~\label{overall framework}
The overall architecture of our TRA is shown in Fig.~\ref{fig:method}.
Following previous works~\citep{HR-Pro, SF-Net,LACP}, we first segment the input video $V = \{I_i\}{_{i=1}^M}$, where $I_i$ denotes the $i$-th frame and $M$ is the total number of video frames, into $T$ multi-frame snippets. A pre-trained video feature extractor (\textit{i.e.}, I3D~\citep{i3d}) is employed to encode these snippets, thereby obtaining snippet-level visual features $\mathbf{F}_{vision}$, which consist of RGB features $\mathbf{F}_{rgb}$ and optical flow features  $ \mathbf{F}_{flow}$.
Meanwhile, we employ a pre-trained image captioner (\textit{i.e.}, BLIP2~\citep{blip2}) to generate a caption for each video frame. The initial captions sequence $\{s_i\}{_{i=1}^M}$ are then fed into our PTR module to obtain improved captions $\{\bar{s}_i\}{_{i=1}^M}$.
%
Given that the frames within a snippet are highly similar to each other, we select the most frequently occurring frame description as the representation for each snippet.
We further encode the snippet-level captions $\{\bar{s}_i\}{_{i=1}^T}$  using a pre-trained textual encoder (\textit{i.e.}, XCLIP~\citep{xclip}) to generate textual features $\mathbf{F}_{text}$.
%
Then, the $\mathbf{F}_{vision}$ and $\mathbf{F}_{text}$ are fed into our PMA module for alignment.
The resulting aligned features are concatenated into $\mathbf{\hat{F}}$, which is subsequently fed into the action detector for temporal action localization. 
We adopt the best-performing HR-Pro~\citep{HR-Pro} as our baseline.

In the following sections, we provide details of our PTR and PMA modules.

\subsection{Point-based Text Refinement}~\label{ptr}
In this section, we explain how the point-based text refinement module enhances the initially generated captions.
\textcolor{black}{First, we categorize actions by their dependency on entities and construct an action-to-entity mapping.} Next, we analyze the flaws in initially generated descriptions, then identify specific video categories and extract the described entities. Subsequently, we build a reliable entity-dependent textual memory based on point annotations. Finally, we replace and remove incorrect entities. We provide details for each step below.

\begin{figure}
\centering
\includegraphics[width=\linewidth]{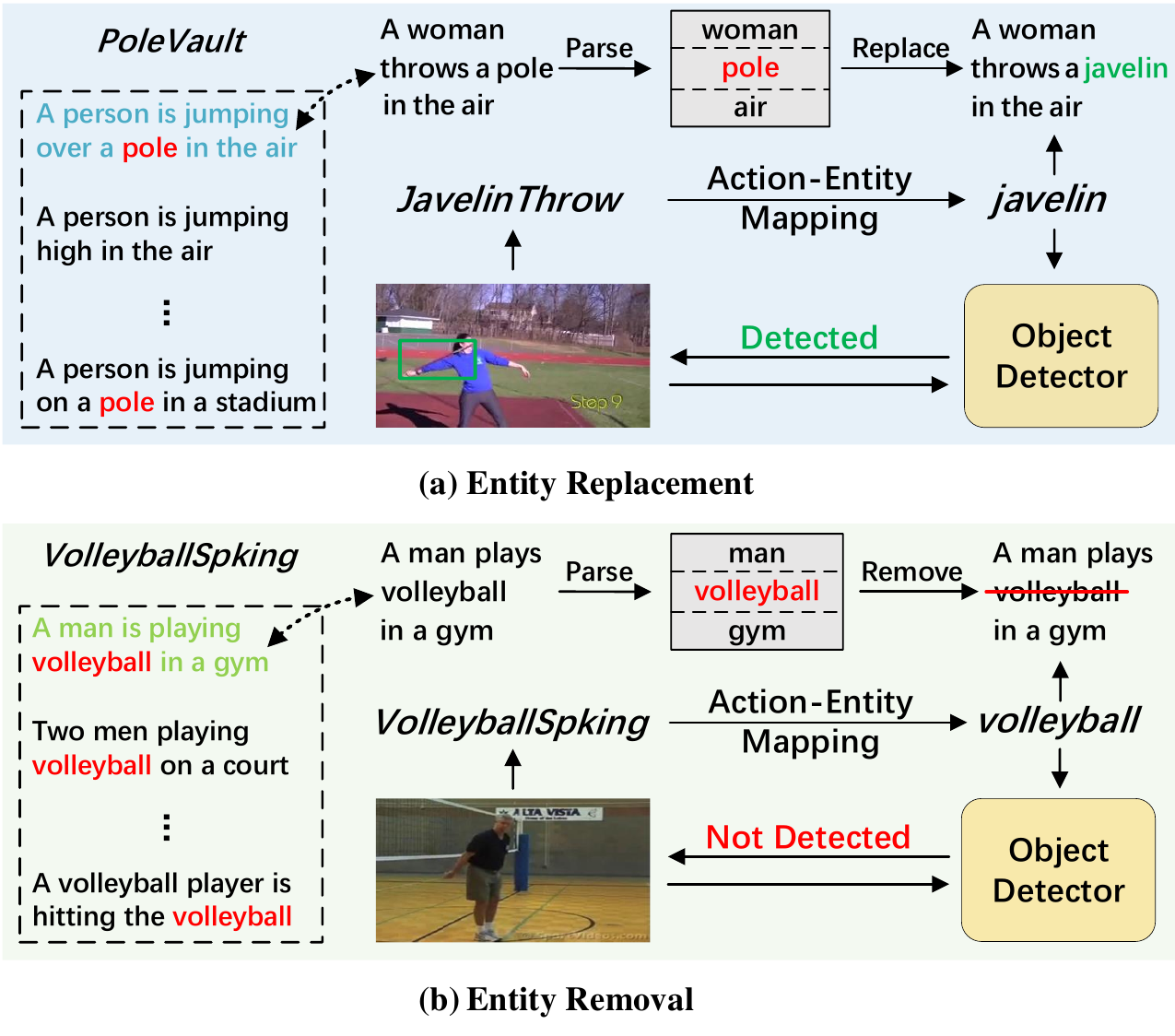}

\caption{
\textcolor{black}{
The proposed PTR module. (a) Entity replacement corrects a mismatched entity in the caption according to the action–entity mapping and object detection. (b) Entity removal deletes an entity that is mentioned in the caption but not detected in the corresponding frame.
}
}
\label{fig:PTR module.}

\end{figure}
\noindent\textbf{Action-to-Entity Mapping Construction.}
We divide actions into two categories: entity-dependent actions and entity-independent actions.
Entity-dependent actions require specific entities to perform, such as discus throw (requires a discus). 
Conversely, entity-independent actions depend mainly on the body's movements and do not need direct involvement with external entities, such as the high jump (which relies primarily on one's own jumping ability).
\textcolor{black}{We identify the most relevant entity for each entity-dependent action, thereby constructing an action-entity mapping $\mathcal{M}:  \{\text{action}_c \to \text{entity}_c\}{_{c=1}^{C_e}} $, where $C_e$ is the number of entity-dependent action categories in the whole dataset.}
The specific details of mapping construction are in Sec.~\ref{Action-Entity Mapping Construction}.

\noindent\textbf{Flaws in Existing Descriptions.}
We observe that due to motion blur and background noise, the descriptions generated from video frames of entity-dependent actions are more likely to generate hallucinations, compared to those from entity-independent actions.
For example, the description in Figure~\ref{fig:PTR module.} (a) incorrectly identifies the ``javelin'' in the video as ``pole''.  
In Figure~\ref{fig:PTR module.}(b), affected by the background, the captioner incorrectly describes ``A man plays volleyball in a gym''.
To minimize entity mistakes in these descriptions, we apply two operations: \textcolor{black}{replacing or removing unreliable entities.}

\noindent\textbf{Video Classification and Entity Parsing.}
For an input video containing entity-dependent actions, its entity-dependent action labels are denoted as $\{y_c\}{_{c=1}^{C_v}}$, where $C_v$ is the number of entity-dependent action categories in the video.
\textcolor{black}{
For training videos, we directly use the annotated video-level labels. For testing videos, the action categories required by PTR are predicted by a separately trained baseline action detector. This baseline detector is independent of the final action detector in TRA and does not share parameters with it. Therefore, no ground-truth action labels or point annotations are used to provide category information to PTR during inference.
}
%
Based on $\mathcal{M}$, we denote the set of entity categories associated with a specific video as $E_v = \{e_c\}{^{C_v}_{c=1}}$, and the complete set of entity categories in the dataset as $E_e$. The complement set $E_x$ contains entities in $E_e$ that do not appear in $E_v$.
%
We use an off-the-shelf textual parser, TextGraphParser~\citep{factualgraph}, to extract the relevant entities from each caption.
Formally, the process is represented as \( s_i \mapsto \{e_{ij}\}_{j=1}^{k_i} \), where \( e_{ij} \) denotes the \( j \)-th entity in the \( i \)-th caption, and \( k_i \) is the total number of entities extracted from that caption.

\noindent\textbf{Build Entity-Dependent Memory.}
We construct a point annotations-based memory as a reference, to evaluate whether a caption describes an action frame.
%
In the point annotation setting, each annotation corresponds to an action category, and a subset of the annotations is related to entity-dependent actions. For these entity-dependent annotations, we also collect the associated descriptions of the annotated frames.
%
\textcolor{black}{
Since some annotated-frame descriptions may contain mismatched entities, as discussed above, we automatically filter these incorrect descriptions using a simple rule-based strategy. Specifically, if the description of an annotated frame contains an entity associated with an irrelevant action category, the description is removed. Since the annotated frame belongs to its labeled action category, entities associated with unrelated action categories are regarded as mismatched. The remaining descriptions are then used to construct a reliable entity-dependent textual memory $\mathbb{M}$, where $\mathbb{M}(y)$ denotes the set of point-level descriptions associated with action category $y$.}

\noindent\textbf{Entity Replacement.}
The generated descriptions with mismatched entities can mislead the model, so we replace them with the correct ones.
If an entity in a sentence belongs to the irrelevant entity set $E_x$, it is necessary to determine whether the sentence describes the action or the background.
Note that background frames may also contain entities in $E_x$. For instance, a discus and a hammer may both be present on the ground, but the player only throws the hammer.
Therefore, irrelevant entities are permitted in background descriptions.
After locating the sentence $s_i$ containing $e_{ij} \in E_x$, we leverage a text embedding model AnglE to calculate the semantic similarity between $s_i$ and each annotated sentence $m$ in $\mathbb{M}(y_{e_{ij}})$.
\begin{flalign}
    & sim(s_i, \mathbb{M}(y_{e_{ij}})) = \max_{ m \in \mathbb{M}(y_{e_{ij}})} \cos(\text{AnglE}(s_i), \text{AnglE}(m))
\end{flalign}
where $\cos$ represents the cosine similarity, $y_{e_{ij}}$ represents the action category of $e_{ij}$.

If $sim(s_i, \mathbb{M}(y_{e_{ij}}))$ exceeds the predefined threshold $\theta_1$, 
$s_i$ requires entity adjustment.
Each $e_v$ in $E_v$ and the frame $I_i$ associated with $s_i$ are jointly fed into the open-set object detector OWLv2~\citep{owlv2}, which predicts the corresponding coordinate $loc_{e_v}$ and confidence score $score_{e_v}$:
\begin{equation}
(loc_{e_v},score_{e_v}) = \textit{Det}(e_{v}, I_{i})   
\end{equation}
where $Det$ represents the entity detection process.
We then identify the entity $\hat e_v$ with the highest confidence:
\begin{equation}
    \hat e_v = \arg\max\limits_{e_v \in E_v}\;score_{e_v}
\end{equation}
Finally, the mismatched entity $e_{ij}$ in sentence $s_i$ is replaced with $\hat e_v$, yielding the updated sentence $\hat{s}_i$.

\noindent\textbf{Entity Removal.}
After fixing mismatched entities, we further remove those entities that appear in the action descriptions but not in the video frames.
Similar to entity replacement, we focus on the action descriptions.
Because the video frame background might contain some title cards, such as ``volleyball from approach and arm swing'', the term ``volleyball'' refers to text, not an actual volleyball, and it should not be deleted.
Specifically, we identify the sentence $\hat{s}_i$ containing $e_{ij} \in E_v$ and use AnglE to compute the semantic similarity between $\hat{s}_i$ and the annotated sentences $m$ in $\mathbb{M}(y_{e_v})$.
\begin{flalign}
    & sim(\hat{s}_i, \mathbb{M}(y_{e_v})) = \max_{ m \in \mathbb{M}(y_{e_{v}})} \cos(\text{AnglE}(\hat{s}_i), \text{AnglE}(m)) 
\end{flalign}

Similar to entity replacement, if $sim(\hat{s}_i, \mathbb{M}(y_{e_v}))$ exceeds the threshold $\theta_2$, $e_{ij}$ and $I_i$ are passed into OWLv2 to generate the coordinates $loc_{e_{ij}}$ and confidence scores $score_{e_{ij}}$:
\begin{equation}
(loc_{e_{ij}},score_{e_{ij}}) = \textit{Det}(e_{ij}, I_{i})   
\end{equation}
If $score_{e_{ij}}$ falls below $\gamma$, $e_{ij}$ is considered undetected and is removed from $\hat{s}_i$; otherwise, no modification is applied.
After removing low-confidence entities (\textit{i.e.}, those with $score_{e_{ij}} < \gamma$),  we obtain the final refined caption, denoted as $\bar{s}_i$.
For brevity, we describe the entity replacement and removal as Algorithm~\ref{alg:ptr}.

\begin{algorithm}[tb]
\caption{Captions Refinement.}
\label{alg:ptr}
\raggedright
\textbf{Input}: Initial captions sequence $\{s_i\}{_{i=1}^M}$,
initial entities sequence $\{\{e_{ij}\}{^{k_i}_{j=1}}\}_{i=1}^M$,
video frames sequence $\{I_i\}{_{i=1}^M}$.
\\
\textbf{Output}: Refined captions sequence $\{\bar{s}_i\}{_{i=1}^M}$.

\begin{algorithmic}[1] 
\FOR{$i=1$ {\bfseries to} $M$}
    \FOR{$j=1$ {\bfseries to} $k_i$}
         \STATE \textbf{if} $e_{ij} \in E_{x}$ \textbf{and} $sim(s_i, \mathbb{M}(y_{e_{ij}})) > \theta_1$  , \textbf{then} $(loc_{e_v},score_{e_v}) =  \textit{Det}(e_{v}, I_{i}), \quad \forall{e_v \in E_v}$ 
         \STATE $\hat e_v = \arg\max\limits_{e_v \in E_v}\;score_{e_v}$ 
         \STATE $ \text{Replace } e_{ij} \text{ with } \hat{e}_v$
         \STATE \textbf{end if}
    \ENDFOR
    \STATE $ \hat{s}_i \gets \{e_{i1}, e_{i2}, \ldots, e_{ik_i}\}$
    \FOR{$j=1$ {\bfseries to} $k_i$}
         \STATE \textbf{if} $e_{ij} \in E_{v}$ \textbf{and}  $  sim(\hat{s}_i, \mathbb{M}(y_{e_v})) > \theta_2$, 
 \textbf{then} $(loc_{e_{ij}},score_{e_{ij}}) = \textit{Det}(e_{ij}, I_{i})$
    \STATE \hspace{1em} \textbf{if} $score_{e_{ij}} < \gamma$, \textbf{then} $ \text{Remove } e_{ij} \text{ from } \hat{s}_i$
    \STATE \hspace{1em} \textbf{end if}
    \STATE \textbf{end if}
    \ENDFOR
    \STATE $\bar{s}_i \gets \{e_{i1}, e_{i2}, \ldots, e_{im_i}\} (m_i<=k_i)$ 
\ENDFOR
\STATE \textbf{Return} $\{\bar{s}_i\}{^{M}_{i=1}}$
\end{algorithmic}

\end{algorithm}

\subsection{Point-based Multimodal Alignment}~\label{pma}
After enhancing the initial captions using PTR, we perform feature alignment between the text features and vision features
(RGB and optical flow). 
Specifically, the three feature types are projected by separate linear layers into a common feature dimension $D$:

\begin{equation}
\begin{aligned}
\hat{\textbf{F}}_{text} &= \text{MLP}_{text}(\textbf{F}_{text}) \\
\hat{\textbf{F}}_{rgb} &= \text{MLP}_{rgb}(\textbf{F}_{rgb}) \\
\hat{\textbf{F}}_{flow} &= \text{MLP}_{flow}(\textbf{F}_{flow})
\end{aligned}
\end{equation}
where $\hat{\textbf{F}}_{text}, \hat{\textbf{F}}_{rgb}$, and $\hat{\textbf{F}}_{flow} \in \mathbb{R}^{T\times D}$.
\textcolor{black}{By leveraging $N_{act}$ pseudo action points and $N_{bkg}$ pseudo background points predicted by the action detector, we select reliable action and background snippets for multimodal contrastive learning. These pseudo action and background points are used to select the corresponding snippets and provide pseudo labels for constructing cross-modal positive and negative pairs.
Specifically, PMA performs bidirectional alignment between textual and visual representations. In the text-to-vision direction, RGB and optical-flow features sharing the same pseudo semantic label as a textual feature are treated as positive samples, whereas those with different pseudo labels serve as negative samples. Symmetrically, in the vision-to-text direction, textual features with the same pseudo label as a visual feature are treated as positive samples, while those with different labels serve as negative samples. 
\textcolor{black}{
The bidirectional contrastive objective therefore enhances cross-modal similarity within the same semantic category and improves the discrimination among different action categories and between action and background representations.
}
}

The text-to-vision contrastive loss $\mathcal{L}_{t2v}$ is calculated as:
\begin{equation}
\begin{aligned}
\mathcal{L}_{t2v} =  - \frac{1}{2N} \sum_{i=1}^N \sum_{j=1}^N \mathbb{I}_{[y_i = y_j]} (\log \frac{\langle\hat{\textbf{f}}_i^{text}, 
\hat{\textbf{f}}_j^{rgb}\rangle}{\sum_{k=1}^N \langle\hat{\textbf{f}}_i^{text}, 
\hat{\textbf{f}}_k^{rgb}\rangle} \\
+ \log \frac{\langle\hat{\textbf{f}}_i^{text}, \hat{\textbf{f}}_j^{flow}\rangle}{\sum_{k=1}^N \langle\hat{\textbf{f}}_i^{text}, \hat{\textbf{f}}_k^{flow}\rangle})
\end{aligned}
\end{equation}
and the vision-to-text contrastive loss $\mathcal{L}_{v2t}$:
\begin{equation}
\begin{aligned}
\mathcal{L}_{v2t} =  - \frac{1}{2N} \sum_{i=1}^N \sum_{j=1}^N \mathbb{I}_{[y_i = y_j]} (\log \frac{\langle\hat{\textbf{f}}_i^{rgb}, 
\hat{\textbf{f}}_j^{text}\rangle}{\sum_{k=1}^N \langle\hat{\textbf{f}}_i^{rgb}, 
\hat{\textbf{f}}_k^{text}\rangle} \\
+ \log \frac{\langle\hat{\textbf{f}}_i^{flow}, \hat{\textbf{f}}_j^{text}\rangle}{\sum_{k=1}^N \langle \hat{\textbf{f}}_i^{flow}, \hat{\textbf{f}}_k^{text} \rangle})
\end{aligned}
\end{equation}
\textcolor{black}{
where $N = N_{act} + N_{bkg}$, and $y_i \in \{0,1,\ldots,C\}$ denotes the pseudo semantic label of the $i$-th selected snippet, with $0$ denoting background and $1,\ldots,C$ denoting the corresponding action categories. Accordingly, $\mathbb{I}_{[y_i=y_j]}$ identifies positive pairs from the same semantic category, whereas samples with different labels, including samples from different action categories and action-background pairs, serve as negatives. $\mathbb{I}$ denotes the indicator function, and $\langle \cdot,\cdot \rangle$ is the similarity function formulated as $\langle\mathbf{x_1},\mathbf{x_2}\rangle = \exp(\mathbf{x_1} \cdot \mathbf{x_2} / \tau)$ with a temperature parameter $\tau$.
}
The total objective function of the PMA module is described as:
\begin{equation}
    \mathcal{L}_{pma} = \frac{1}{2}\mathcal{L}_{t2v} + \frac{1}{2}\mathcal{L}_{v2t}.
\end{equation}

\textcolor{black}{Notably, pseudo action/background point mining may introduce noisy samples and result in incorrect cross-modal pairings. Nevertheless, reliable pseudo points provide additional training samples, promote cross-modal consistency within the same semantic category, and improve the discrimination among different action categories and between action and background representations.}

\subsection{Training}~\label{training}
\textcolor{black}{
The pretrained components, including BLIP2, TextGraphParser, OWLv2, XCLIP, and AnglE, are used as fixed preprocessing or feature extraction modules, and their parameters are not updated during training. Only the trainable localization network and the proposed alignment module are optimized with the total loss:
}
\begin{equation}
    \mathcal{L}_{total} = \mathcal{L}_{base} + \lambda~\mathcal{L}_{pma},
\end{equation}
where $\mathcal{L}_{base}$ represents the original loss function of the base action detector, $\lambda$ is a trade-off hyperparameter. More details of $\mathcal{L}_{base}$ are given in Sec.~\ref{base_loss}.

\subsection{Inference}~\label{inference}
\textcolor{black}{
At test time, we first use a separately trained baseline action detector to predict the video-level action categories required by PTR. PTR then refines the initial captions according to these predicted categories. The resulting textual features, together with the RGB and optical-flow features, are subsequently projected into a common feature space using the learned projection layers of PMA.
 The aligned features are concatenated as $\hat{\textbf{F}} = \textit{Concat}(\hat{\textbf{F}}_{rgb}, \hat{\textbf{F}}_{flow}, \hat{\textbf{F}}_{text})$ and then fed into the final action detector of TRA to generate a series of proposals. Finally, we use Soft-NMS~\citep{soft_nms} to remove overlapping proposals. The baseline detector used for category prediction and the final action detector are separately trained models and do not share parameters.
}

\section{Experiment}

\subsection{Experimental Setup}

\noindent\textbf{Dataset}. In this work, we conduct experiments on five widely used PTAL benchmarks.
\textbf{THUMOS'14}~\citep{th14} is a widely used challenging dataset for Temporal Action Localization. It contains 413 untrimmed sports videos, covering 20 action categories with an average of 15 action instances per video.  Following standard divisions, we use 200 videos for training and 213 videos for testing.
\textbf{BEOID}~\citep{beoid} provides 58 video samples with 34 operation classes in 6 locations, including 46 videos for training and 12 videos for testing. 
\textbf{GTEA}~\citep{gtea} includes 28 untrimmed videos of 7 fine-grained daily activities in a kitchen.  We adopt the standard split, which involves using 21 videos for training, and 7 videos for testing.
\textbf{ActivityNet}~\citep{activitynet} is a large-scale dataset designed for action detection and offers two different versions. Version 1.3 includes 10,024 training videos, 4,926 validation videos, and 5,044 test videos, divided into 200 action categories. Version 1.2, a subset of version 1.3, consists of 4,819 training videos, 2,383 validation videos, and 2,480 test videos, covering 100 action categories. We conduct experiments on both versions of ActivityNet to validate the generalization of our approach.

\noindent\textbf{Evaluation Metrics}.
We follow the standard protocol to evaluate the action localization performance across three datasets using mean Average Precision (mAP) at different Intersection-over-Union (IoU) thresholds. A proposal is considered positive only if it exceeds the specified IoU threshold and the correct category prediction.

\noindent\textbf{Implementation Details}.
Following~\citep{LACP,SF-Net,HR-Pro}, we split each video into non-overlapping 16-frame snippets and use the TV-L1 algorithm to extract optical flow.
For a fair comparison, we also follow the previous works to use a two-stream I3D~\citep{i3d} network pre-trained on Kinetics-400 to extract the visual features.
The sentence similarity threshold $\theta_1$ and $\theta_2$ are set to 0.4 and 0.75, respectively.
The detection threshold of OWLv2 is set to 0.1. 
In the training, the balance factor $\lambda$ is set to 0.9, the temperature parameter $\tau$ is set to 1.0, the learning rate is set to 0.0002, and other parameters (\textit{e.g.}, batch size) are maintained as in the baseline.
\textcolor{black}{At test time, the baseline detector used to provide video-level action-category predictions for PTR is the original HR-Pro model without the proposed PTR and PMA modules. It is trained separately from the final TRA model, and the two detectors do not share parameters. Only the categories predicted by this baseline detector are used for PTR, without access to ground-truth action labels or point annotations.
}
Our entire framework is implemented with PyTorch and all experiments are conducted on 1 $\times$ NVIDIA GeForce RTX-3090 GPU.

\noindent\textbf{Specific Pre-trained Models}
We use several pre-trained models throughout the pipeline to achieve our goal.
Specifically,
We use the pre-trained BLIP2\footnote{https://huggingface.co/Salesforce/blip2-opt-6.7b}~\citep{blip2} as our caption generator, which is fine-tuned on web-scale image-text datasets (\textit{e.g.}, LAION~\citep{laion}) and has strong zero-shot image captioning abilities.
The pre-trained XCLIP\footnote{https://huggingface.co/microsoft/xclip-base-patch16-16-frames}~\citep{xclip} is employed to extract the text features, which is pre-trained on large-scale action recognition Kinetics-400~\citep{k400} dataset and provides powerful representation for action-related sentences.
%
%
The pre-trained TextGraphParser\footnote{https://huggingface.co/lizhuang144/flan-t5-base-VG-factual-sg}~\citep{factualgraph} serves as the sentence parser to extract entities,  which is fine-tuned on FACTUAL~\citep{factualgraph} scene graph parsing dataset and can capture the accuracy entities from the given sentence.
We leverage the OWLv2\footnote{https://huggingface.co/google/owlv2-large-patch14-ensemble}~\citep{owlv2} to detect if the key entities are in the video frames, which uses CLIP~\citep{clip} as its multi-modal backbone and has the excellent zero-shot text-conditioned object detection ability.
AngLE\footnote{https://huggingface.co/WhereIsAI/UAE-Large-V1}~\citep{angle} demonstrates state-of-the-art (SOTA) performance on semantic textual similarity tasks, so we use it to compute the similarity between the generated caption and entity-dependent memory.
The aforementioned pre-trained models are publicly and readily available.

\begin{figure}
\centering
\includegraphics[width=\linewidth]{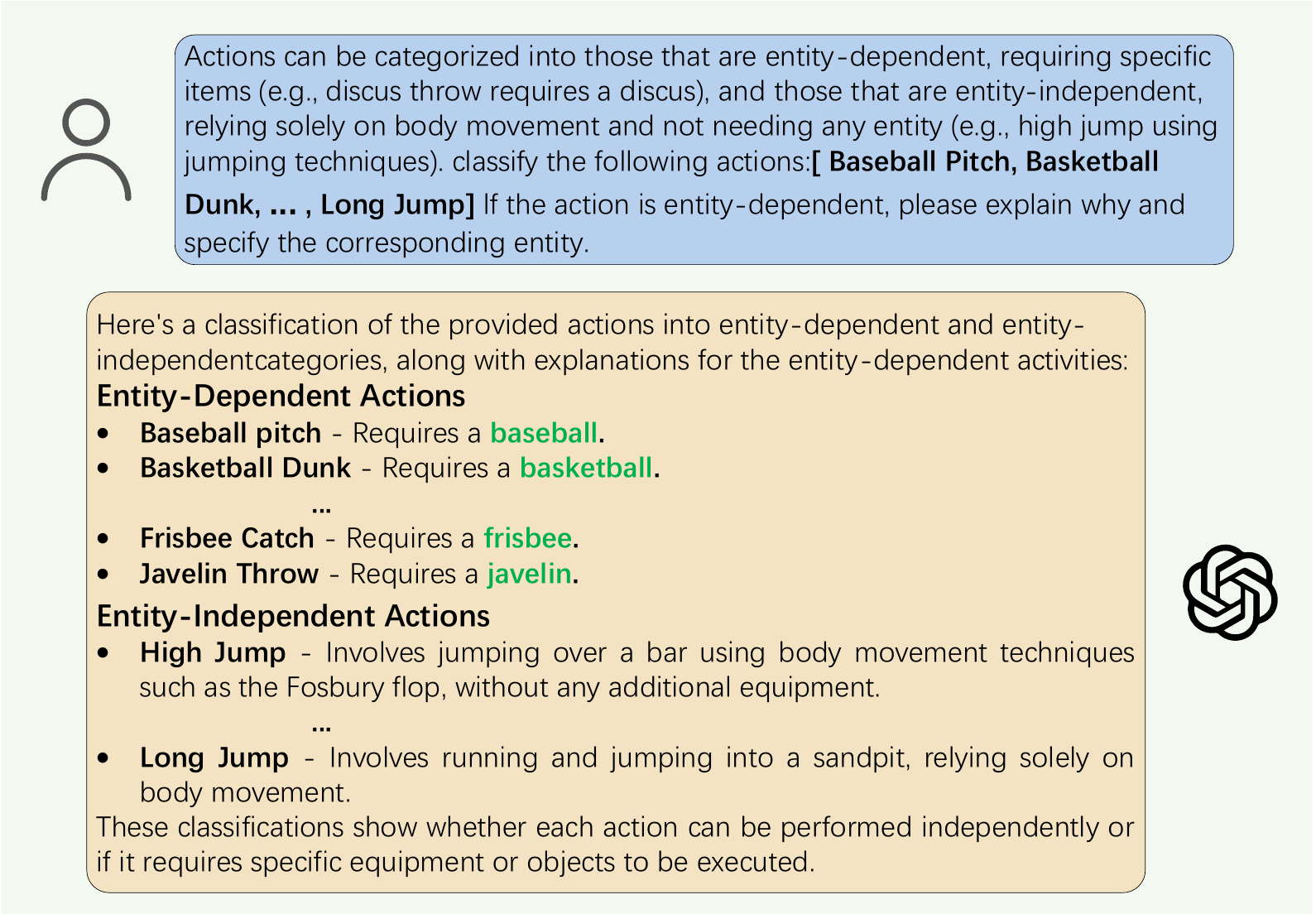}
\caption{
\textcolor{black}{
Using ChatGPT-4 to generate the initial action–entity mapping by classifying actions as entity-dependent or entity-independent and identifying the corresponding entities, followed by manual verification and correction.
}
}
\label{fig:chatgpt}
\end{figure}

\noindent\textbf{Action-Entity Mapping Construction}~\label{Action-Entity Mapping Construction}
The construction of action-entity mapping consists of two steps: (1) Using a large language model to automatically classify action categories and identify related entities according to the predefined standards in Sec.~\ref{ptr}, and (2) Manually adjusting the above results to guarantee accuracy.
%
In detail, we use ChatGPT-4\footnote{https://openai.com} to build the action-entity mapping, as shown in Fig.~\ref{fig:chatgpt}.
Due to the hallucination and randomness inherent in large language models, we manually adjust the initially generated results.
The final action-entity mapping we used is shown in Table~\ref{action-entity-mapping}.
\begin{table}[H]
    \centering
    \caption{The specific action-entity mapping. The action categories are divided into entity-dependent and entity-independent actions. Each entity-dependent action is paired with its most relevant entity, while entity-independent actions are marked with ``--''.}
    \resizebox{0.8\linewidth}{!}{
        \setlength{\tabcolsep}{1mm}{
            \begin{tabular}{c|c|c}
                \toprule
                \multicolumn{3}{c}{Action-Entity Mapping} \\
                \midrule
                Number&Action&Entity \\
                \midrule
                1&Baseball Pitch&baseball \\
                2&Basketball Dunk&basketball \\
                3&Billiards&billiard ball \\
                4&Clean and Jerk&barbell \\
                5&Cricket Bowling&cricket ball \\
                6&Cricket Shot&cricket bat \\
                7&Golf Swing&golf club \\
                8&HammerThrow&hammer \\
                9&Pole Vault&pole \\
                10&Soccer Penalty&soccer ball \\
                11&Tennis Swing&tennis racket \\
                12&Shot Put&shot \\
                13&Throw Discus&discus \\
                14&Volleyball Spiking&volleyball \\
                15&Frisbee Catch&frisbee \\
                16&Javelin Throw&javelin \\
                17&High Jump&- \\
                18&Cliff Diving&- \\
                19&Diving&- \\
                20&Long Jump&- \\

                \bottomrule
            \end{tabular}            
        }
    }
     \vspace{-1mm}
     \label{action-entity-mapping}
\end{table}%

\noindent\textbf{Base Loss of Action Detector}~\label{base_loss}
In Sec.~\ref{training}, we use the original loss function of the baseline as $\mathcal{L}_{base}$.
Taking the baseline HR-Pro~\citep{HR-Pro} as an example, $\mathcal{L}_{base}$ consists of the frame-wise loss function $\mathcal{L}_{frame}$ and the feature-wise loss function $\mathcal{L}_{feature}$:
\begin{equation}
{\mathcal L}_{base} = {\mathcal L}_{frame} + {\mathcal L}_{feature}
\end{equation}
where
\begin{equation}
\begin{small}
{\mathcal L}_{frame}=\frac{1}{N_{act}} \sum_{c=1}^{C} \sum_{t\in{\mathcal{T}^+}}^{N_{act}}FL(\mathbf{P}_{t,c})
 + \frac{1}{N_{bkg}} \sum_{t\in{\mathcal{T}^-}}^{N_{bkg}} FL(1-\mathbf{A}_{t})
\label{eq_4}
\end{small}
\end{equation}
  $N_{act}$ and $N_{bkg}$ is the total number of pseudo action snippets $\mathcal{T}^+$ and background snippets $\mathcal{T}^-$ respectively,  $\mathbf{P}$ and $\mathbf{A}$ are the snippet-level prediction and class-agnostic attention sequence, $C$ represents the number of total action classes, \textit{FL} represents the focal loss function~\citep{focalloss}.

\begin{equation}
\begin{small}
\begin{aligned}
{\mathcal L}_{feature} = -\frac{1}{C}\sum_{c=1}^{C}\sum_{t_i^c}log(
                        \frac{\langle\bar{\mathbf{x}}_{t_i^c} \cdot \bar{\mathbf{m}}_c \rangle}
                        {\langle\bar{\mathbf{x}}_{t_i^c} \cdot \bar{\mathbf{m}}_c \rangle+\sum_{\forall k \neq c}\langle\bar{\mathbf{x}}_{t_i^c} \cdot \bar{\mathbf{m}}_k \rangle} \\
                        +\frac{\langle\bar{\mathbf{x}}_{t_i^c} \cdot \bar{\mathbf{m}}_c \rangle}
                        {\langle\bar{\mathbf{x}}_{t_i^c} \cdot \bar{\mathbf{m}}_c \rangle+\sum_{\forall t_j \in \mathcal{T}^-}\langle\bar{\mathbf{x}}_{t_j} \cdot \bar{\mathbf{m}}_c \rangle})
\end{aligned}
\end{small}
\end{equation}
where $t_i^c$ indicates the pseudo action snippet of class c, $\bar{\mathbf{x}}$ represent the normalized embedded snippet features, $\bar{\mathbf{m}}_c$ represent the normalized prototype of class c, $\langle \cdot,\cdot \rangle$ is the same similarity operation in Sec.~\ref{training}.

\begin{table*}[t]
\centering
\caption{
Comparison with the state-of-the-art methods on THUMOS'14.
We also compare the method with fully-supervised and weakly-supervised methods.
We use the same point annotations as ~\citep{LACP}. $\uparrow$ highlights the absolute performance improvement between the best and second-best methods in point-level supervision. To compare fairly with proposal-based refinement methods~\citep{TSPNet, HR-Pro, dcm}, we also employ a proposal-based boundary adjustment, similar to~\citep{HR-Pro}. $ ^{*}$ represents our proposal-based results.} 
\resizebox{\textwidth}{!}{
\begin{tabular}{c|l|ccccccc|ccc}
\hline\hline

\multirow{2}{*}{Supervision}       & \multicolumn{1}{c|}{\multirow{2}{*}{Method}} &
\multicolumn{7}{c|}{mAP@IoU (\%)}  & AVG  & AVG  & AVG  \\
    & \multicolumn{1}{c|}{}  & 0.1  & 0.2  & 0.3  & 0.4  & 0.5  & 0.6  & 0.7 & (0.1:0.5)  & (0.3:0.7) & (0.1:0.7)    \\
    \hline\hline
    
\multirow{5}{*}{\begin{tabular}{c}Frame-level\\(Full)\end{tabular}}
    & P-GCN~\citep{p-gcn} \small {(ICCV'19)}  & 69.5  & 67.8  & 63.6  & 57.8  & 49.1  & -  & -  & 61.6  & - & - \\
    & TCANet~\citep{tcanet} \small {(CVPR'21)}  & -  & -  & 60.6  & 53.2  & 44.6  & 36.8  & 26.7  & -  & 44.3 & -\\
    & BMN+CSA~\citep{BMN+CSA} \small {(ICCV'21)} & -  & -  & 64.4  &58.0  & 49.2 & 38.2  & 27.8  & -  & 47.5 & -\\
     & GCM~\citep{gcm} \small {(TPAMI'21)}  & 72.5  & 70.9  & 66.5  & 60.8  & 51.9  & -  & -  & 64.5  & - & -\\
     & VSGN~\citep{vsgn} \small {(ICCV'21)} & -  & -  & 66.7  &60.4  & 52.4 & 41.0  & 30.4  & -  & 50.2 & -\\
    & AFSD~\citep{AFSD} \small {(CVPR'21)} & -  & -  & 67.3  & 62.4  & 55.5  & 43.7  & 31.1  & -  & 52.0 & -\\
    
    \hline\hline
    
\multirow{6}{*}{\begin{tabular}{c}Video-level\\(Weak)\end{tabular}}
    & ACG-Net~\citep{acg-net} \small{(AAAI'22)} & 68.1  & 62.6  & 53.1  & 44.6  & 34.7  & 22.6  & 12.0  & 52.6  & 33.4 & 42.5\\
    & RSKP~\citep{RSPK} \small {(CVPR'22)}   & 71.3  & 65.3  & 55.8  & 47.5  & 38.2  & 25.4  & 12.5  & 55.6  & 35.9  & 45.1\\
    & DELU~\citep{delu} \small {(ECCV'22)}   & 71.5  & 66.2  & 56.5  & 47.7  & 40.5  & 27.2  & {15.3}  & 56.5  & 37.4  & 46.4\\    
    & P-MIL~\citep{P-MIL} \small {(CVPR'23)} & 71.8  & 67.5  & 58.9  &49.0  & 40.0 & 27.1  & 15.1  & 57.4  & 38.0 & 47.0\\
    & Zhou~\etal~\citep{zhou_et_al} \small {(CVPR'23)}   & 74.0  & 69.4  & 60.7  & 51.8  & 42.7  & 26.2  & 13.1  & 59.7  &38.9  & 48.3\\
    & PivoTAL~\citep{pivotal} \small {(CVPR'23)} & 74.1  & 69.6  & 61.7  &52.1  & 42.8 & 30.6  & 16.7  & 60.1  & 40.8 & 49.6\\
    & ISSF~\citep{ISSF} \small{(AAAI'24)} &72.4 &66.9 &58.4 &49.7 &41.8 &25.5 &12.8 &57.8 &37.6 &46.8 \\
    &PVLR~\citep{PVLR}  \small{(MM'24)} &74.9 &69.9 &61.4 &53.1 &45.1 &30.5 &17.1 &60.9&-&50.3 \\

    \hline
    \multirow{10}{*}{\begin{tabular}{c}Point-level\\(Weak)\end{tabular}}
    &ARST~\citep{ARST} \small {(CVPR'19)} &24.3 &19.9 &15.9 &12.5 &9.0 &- &- &16.30 &- &- \\
    & SF-Net~\citep{SF-Net} \small {(ECCV'20)}  & 68.3  & 62.3  & 52.8  & 42.2  & 30.5  & 20.6  & 12.0   & 51.2  & 31.6  & 41.2\\
    & LACP~\citep{LACP} \small {(ICCV'21)}  & 75.7  & 71.4  & 64.6  & 56.5  & 45.3  & 34.5  &  21.8  & 62.7  &  44.5  & 52.8\\
    & BackTAL~\citep{backtal} \small {(TPAMI'21)} &-&-&54.4&45.5&36.3&26.2&14.8&-&35.4&- \\
    & CRRC-Net~\citep{crrc-net} \small{(TIP'22)} & 77.8  & 73.5  & 67.1  & 57.9  & 46.6  & 33.7  &  19.8  & 64.6  &  45.1  & 53.8\\
    & SMBD~\citep{SMBD} \small{(ECCV'24)} & -  & -  & 66.0  & 57.9  & 47.0  & 36.0  &  22.0  & 64.2  &  45.7  & -\\
    & SNPR~\citep{snpr} \small{(TIP'24)} & 77.9  & 73.9  & 66.6  & 59.4  & 48.6  & 36.7  &  22.7  & 65.3  &  46.8  & 55.1\\
    & AAPL~\citep{AAPL} \small{(AAAI'25)} & 64.3  & -  & 54.6  & -  & 35.2  & -  &  14.0  & -  &  -  & 42.8\\
    &  $\textbf{TRA}~{\textbf{(Ours)}}$
  & \textbf{83.6}  & \textbf{79.4}  & \textbf{71.6}  & \textbf{62.0}  & \textbf{51.4}  & \textbf{37.6}  & \textbf{23.7}  & \textbf{69.6}  &  \textbf{49.3} &\textbf{58.5} \\
    \cline{2-12}
    & DCM~\citep{dcm} \small {(ICCV'21)}  & 72.3  & 64.7  & 58.2  & 47.1  & 35.9  & 23.0  & 12.8   & 55.6  & 35.4  & 44.9\\
    &  TSPNet~\citep{TSPNet} \small {(CVPR'24)}  & 82.3  & 77.6  & 70.1  & 60.0  & 49.4  & 37.6  & 24.5  & 67.9  &  48.3 &57.4 \\
    &  HR-Pro~\citep{HR-Pro} \small {(AAAI'24)}  & 85.6  & 81.6  & 74.3  & 64.3  & 52.2  & 39.8  & 24.8  & 71.6  &  51.1 &60.3 \\
    &  QROT~\citep{QROT} \small {(CVPR'25)}  & -  & -  & 73.1  & 64.4  & 54.3  & 41.3  & \textbf{27.4}  & 72.3  &  52.1 &- \\
    &  $\textbf{TRA}^{*}~{\textbf{(Ours)}}$
  & \textbf{86.6}  & \textbf{83.6}  & \textbf{76.3}  & \textbf{66.7}  &\textbf{55.2}  & \textbf{41.4}  & {26.5}  & \textbf{73.7}  &  \textbf{53.2} &\textbf{62.3} \\
    
    \hline\hline
    
\end{tabular}
}
\label{table:thumos_benchmark}
\end{table*}

\begin{table*}[t]
\centering
\caption{
Comparison results with the state-of-the-art methods on GTEA and BEOID. $\uparrow$ represents the absolute gain between the best and second-best PTAL methods.  We make the same proposal-based boundary adjustment~\citep{HR-Pro} for a fair comparison with two-stage methods~\citep{TSPNet, HR-Pro, dcm}.   $ ^{*}$ denotes our refinement results.}
\footnotesize
\setlength{\tabcolsep}{4pt}
\begin{tabular}{c|l|cccc|c}
\hline\hline

\multirow{2}{*}{Dataset}       & \multicolumn{1}{c|}{\multirow{2}{*}{Method}} &
\multicolumn{4}{c|}{mAP@IoU (\%)}    & AVG  \\
    & \multicolumn{1}{c|}{}  & 0.1   & 0.3   & 0.5  & 0.7  & (0.1:0.7)    \\
    \hline\hline
    
\multirow{9}{*}{GTEA}
    & SF-Net~\citep{SF-Net} \small {(ECCV'20)}  & 58.0  & 37.9   & 19.3  & 11.9 & 31.0 \\
    & LACP~\citep{LACP} \small {(ICCV'21)}  & 63.9  & 55.7   & 33.9  & 20.8 & 43.5 \\
    & SMBD~\citep{SMBD} \small {(ECCV'24)}  & 75.0  & 61.3   & 41.1  & 14.2 & 47.4 \\
    & SNPR~\citep{snpr} \small {(TIP'24)}  & 74.3  & 62.8   & 35.7  & 13.7 & 46.6 \\
    & AAPL~\citep{AAPL} \small(AAAI'25)  & 70.3  & 54.4  & 37.7 & 23.4 & 46.3 \\
    & \textbf{TRA}   & \textbf{76.0}  & \textbf{66.0}   & \textbf{46.8}  & 
 \textbf{22.0} & \textbf{52.8} \\
    \cline{2-7}
    & DCM~\citep{dcm} \small {(ICCV'21)}  & 59.7  & 38.3   & 21.9  & 18.1 & 33.7 \\
    & TSPNet~\citep{TSPNet} \small {(CVPR'24)}  & 74.6  & 60.9   & 39.5  & 16.6 & 49.0 \\
    & HR-Pro~\citep{HR-Pro} \small {(AAAI'24)}  & 72.6  & 61.1   & 37.3  & 17.5 & 47.3 \\
    & $\textbf{TRA}^{*}~{\textbf{(Ours)}}$   & \textbf{77.9}  & \textbf{67.8}   & \textbf{47.2}  & \textbf{22.4} & \textbf{54.1} \\
    \hline\hline
    
\multirow{9}{*}{BEOID}
    & SF-Net~\citep{SF-Net} \small {(ECCV'20)}  & 62.9  & 40.6   & 16.7  & 3.5 & 30.9 \\
    & LACP~\citep{LACP} \small {(ICCV'21)}  & 76.9  & 61.4   & 42.7  & 25.1 & 51.8 \\
    & BackTAL~\citep{backtal} \small {(TPAMI'21)} &60.1 &40.9 &21.2 &11.0 &32.5 \\
    & SMBD~\citep{SMBD} \small {(ECCV'24)}  & 78.2  & 71.0   & 52.5  & 25.2 & 57.4 \\
    & SNPR~\citep{snpr} \small {(TIP'24)}  & 77.2  & 64.3   & 44.0  & 24.5 & 53.1 \\
    & AAPL~\citep{AAPL} \small(AAAI'25) & 75.5  & 67.6  & 48.5 & 26.3 & 55.2 \\
    & \textbf{TRA}   & \textbf{82.1}  & \textbf{71.0}   & \textbf{55.1}  & \textbf{30.0} & \textbf{60.8} \\
    \cline{2-7}
    & DCM~\citep{dcm} \small {(ICCV'21)}  & 63.2  & 46.8   & 20.9  & 5.8 & 34.9 \\
    & TSPNet~\citep{TSPNet} \small {(CVPR'24)}  & \textbf{83.8}  & 73.0   & 51.1  & 23.8 & 59.6 \\
    & HR-Pro~\citep{HR-Pro} \small {(AAAI'24)}  & 78.5  & 72.1   & 55.3  & 26.1 & 59.4 \\
    & $\textbf{TRA}^{*}~{\textbf{(Ours)}}$   & 83.3  & \textbf{74.5}   & \textbf{58.9}  & \textbf{31.4} & \textbf{63.7} \\
    \hline\hline
    
\end{tabular}
\label{table:gtea_beoid_benchmark}
\end{table*}

\begin{table*}[t]
\centering
\caption{
Comparison results with the state-of-the-art methods on ActivityNet1.2 and ActivityNet1.3. $ ^{*}$ denotes our boundary adjustment results, identical to the proposal-based refinements in~\citep{HR-Pro}.
}
\footnotesize
\setlength{\tabcolsep}{4pt}
\begin{tabular}{c|l|ccc|c}
\hline\hline

\multirow{2}{*}{Dataset}       & \multicolumn{1}{c|}{\multirow{2}{*}{Method}} &
\multicolumn{3}{c|}{mAP@IoU (\%)}    & AVG  \\
    & \multicolumn{1}{c|}{}  & 0.5   & 0.75   & 0.95  & (0.5:0.95)    \\
    \hline\hline
    
\multirow{5}{*}{ActivityNet1.2}
    & SF-Net~\citep{SF-Net} \small {(ECCV'20)}  & 37.8     & -  &- & 22.8 \\
    & LACP~\citep{LACP} \small {(ICCV'21)}  & 44.0    & 26.0  & 5.9 & 26.1 \\
    & BackTAL~\citep{backtal} \small {(TPAMI'21)} &41.5 &27.3 &4.7 &27.0\\
    & SNPR~\citep{snpr} \small {(TIP'24)}  & 43.4     & \textbf{31.3}  & 5.4 & 27.5 \\
    & \textbf{TRA}   & \textbf{46.8}  & {27.3}    & 
 \textbf{6.7} & \textbf{28.4} \\
    \cline{2-6}
    & $\textbf{TRA}^{*}~{\textbf{(Ours)}}$   & \textbf{49.6}  & \textbf{29.0}     & \textbf{7.3} & \textbf{30.2} \\
    \hline\hline
    
\multirow{6}{*}{ActivityNet1.3}
    & LACP~\citep{LACP} \small {(ICCV'21)}  & 40.4     & 24.6  & 5.7 & 25.1 \\
    & CRRC~\citep{crrc-net} \small {(TIP'22)}  & 39.8     & 24.1  & 5.9 & 24.0 \\
    & SNPR~\citep{snpr} \small {(TIP'24)}  & 41.3     & \textbf{30.9}  & 4.8 & 26.5 \\
    & AAPL~\citep{AAPL} \small(AAAI'25)  & 39.6  & 24.3  & 5.6 & 24.7 \\
    & \textbf{TRA}   & \textbf{43.9}  & {26.4}   & \textbf{6.5}   & \textbf{27.1} \\
    \cline{2-6}
    & HR-Pro~\citep{HR-Pro} \small {(AAAI'24)}  & 42.8    & 27.2  & \textbf{8.0} & 27.1 \\
    & $\textbf{TRA}^{*}~{\textbf{(Ours)}}$   & \textbf{45.3}  & \textbf{27.4}  & 6.8   & \textbf{28.0} \\
    \hline\hline   
\end{tabular}
\label{table:activitynet1.2_1.3_benchmark}
\end{table*}

\subsection{Comparison with State-of-The-Art Methods}
 In Table~\ref{table:thumos_benchmark}, we evaluate our TRA against several state-of-the-art methods under different levels of supervision on THUMOS'14.
 Our model achieves the best performance among all point-level and video-level weakly-supervised approaches.
 In snippet-level approaches, our method outperforms the second-ranked SNPR, achieving absolute improvements of 4.3\%, 2.5\%, and 3.4\% on the metrics of average mAP (0.1:0.5), average mAP (0.3:0.7), and average mAP (0.1:0.7) respectively.
 %
 Compared to PTAL methods that perform boundary adjustment in the second phase, our proposal-based refinement TRA reaches an average mAP (0.1:0.7) of 62.3\%, leading by about 2.0\% over HR-Pro.
 It also surpasses the recent QROT by 1.4 and 1.1 percentage points in average mAP@[0.1:0.5] and mAP@[0.3:0.7], respectively.
 When compared with some fully supervision methods, such as AFSD, our method still shows a superior performance (53.2\% \textit{v.s.} 52.0\% in average mAP for IoU thresholds of 0.3:0.7).
Table~\ref{table:gtea_beoid_benchmark} demonstrates the impressive performance of our method on the GTEA and BEOID benchmarks.
In both datasets, our method gains an absolute improvement of 5.4\% and 3.4\% in terms of average mAP (0.1:0.7), while the proposal-based refinement surpasses comparable methods by 5.1\% and 4.1\%.
Given that both datasets consist of ego-centric daily activities rather than complex sports actions, we remove the PTR module.
%
We also evaluate our method on large-scale action detection datasets.
As shown in Table~\ref{table:activitynet1.2_1.3_benchmark}, our method achieves the best performance on both the ActivityNet 1.2 and ActivityNet 1.3 large-scale datasets, with 30.2\% mAP and 28.0\% mAP, respectively.

\subsection{Ablation Study}
In this section, we perform a set of ablation studies on THUMOS'14 as it is the most challenging.
\begin{table}[H]
\centering
\caption{
Ablation study on THUMOS14.
$\uparrow$ denotes the relative gain between our full model and baseline.
}
\resizebox{0.5\textwidth}{!}{\setlength{\tabcolsep}{1mm}{
\begin{tabular}{ccc|ccc}
    \hline\hline
    \multicolumn{3}{c|}{{Setup}} 
    & \multicolumn{3}{c}{AVG mAP@IoU(\%)} \\
    \hline
    TEXT & PTR & PMA  
      & [0.1:0.5] & [0.3:0.7] & [0.1:0.7] \\
    \hline
      \XSolidBrush& \XSolidBrush  & \XSolidBrush    & 67.4 & 46.9     & 56.2  \\
    \Checkmark & \XSolidBrush  &  \XSolidBrush   & 68.0 & 47.5    & 56.9   \\
    \Checkmark & \Checkmark  &  \XSolidBrush  &68.8 & 48.2    & 57.6 \\
    \Checkmark & \Checkmark  & \Checkmark   &\textbf{69.6}{$^{\uparrow2.2}$} & \textbf{49.3}{$^{\uparrow2.4}$}    & \textbf{58.5}{$^{\uparrow2.3}$}  \\
    \hline\hline
\end{tabular}
}
}
\label{table:ablation_studies}
\end{table}

\noindent\textbf{Effectiveness of Each Component.}
Table~\ref{table:ablation_studies} presents the comparison results of using different modules of TRA. 
The first line represents the baseline, which uses only visual features without any extra modules, achieving an average mAP (0.1:0.7) of 56.2\%.
Incorporating the initial text modality generated by BLIP2 leads to a 0.7\% increase in average mAP for IoU thresholds of 0.1:0.7 (Row 2 \textit{v.s.}~Row 1).
By augmenting the original caption with the proposed PTR, the model realizes a further increase in performance,  reaching an average mAP (0.1:0.7) of 57.6\% (Row 3).
\textcolor{black}{Following the alignment of visual and textual modalities through PMA, our full model achieves the best average mAP (0.1:0.7) with a score of 58.5\%. Compared with the TEXT+PTR setting, PMA brings a further improvement of 0.9\% (Row 4 \textit{v.s.}~Row 3), and the full model surpasses the visual-only baseline by 2.3\% (Row 4 \textit{v.s.}~Row 1).
}
This clearly shows that our proposed modules enhance and complement precise action localization.

\noindent\textbf{Effect of Different Base Action Locators.}
To validate the generalization of our method, we plug our proposed module into two different state-of-the-art base action locators.
The results in Table~\ref{table:Generalization} demonstrate
that our method achieves improvements across different baselines, indicating its generalizability.
We observe that our method brings more performance improvements to HR-Pro compared to LACP.
This is because a stronger baseline generates more accurate pseudo action points and background points, resulting in greater enhancements to the PMA.
\begin{table}[H]
    \centering
    \caption{Results of incorporating TRA into different PTAL methods on THUMOS’14. $^\ddagger$ represents our reproduced results.}
    \resizebox{\linewidth}{!}{
        \setlength{\tabcolsep}{1mm}{
            \begin{tabular}{l|ccccc}
                \toprule
                \multicolumn{1}{c|}{\multirow{2}{*}{Method}}  & \multicolumn{5}{c}{mAP@IoU(\%)} \\
                \cline{2-6}
                  & 0.1 & 0.3 & 0.5 & 0.7 &AVG \\
                \midrule
                LACP$^\ddagger$  & 75.3 & 64.3 & 45.1&20.6 &52.4 \\
                LACP + Ours  & 79.1 & 67.8 & 45.4&20.8 &54.1{$^{\uparrow1.7}$} \\
                HR-Pro$_{snippet}^\ddagger$  & 82.0 & 69.4 & 48.2&21.9 &56.2 \\
                HR-Pro$_{snippet}$ + Ours  & 83.6 & 71.6 & 51.4&23.7 &58.5{$^{\uparrow2.3}$} \\
                \bottomrule
            \end{tabular}            
        }
    }
     \label{table:Generalization}
\end{table}%

\noindent\textbf{Effect of Different Caption Feature Extractor.}
In Table~\ref{caption feature extractor}, we compare the performance of our method under different caption feature extractors.
We observe that VIT-B/16 outperforms VIT-B/32 when used as the backbone~(Row~1 \textit{v.s.} Row~3 \& Row~2 \textit{v.s.} Row~4), primarily because it can capture more fine-grained features.
We also notice that fine-tuning on the Kinetics-400 dataset leads to better performance, as it is a large action recognition dataset.
Overall, using XCLIP-VIT B/16 as the text feature extractor achieves superior performance.

\begin{table}[H]
    \centering
    \caption{Performance with different Caption Feature Extractor.}
    \resizebox{\linewidth}{!}{
        \setlength{\tabcolsep}{1mm}{
            \begin{tabular}{l|l|c|ccc}
                \toprule
                \multicolumn{1}{c|}{\multirow{2}{*}{Model}} & \multirow{2}{*}{Pre-trained} & \multirow{2}{*}{Fine-tune} & \multicolumn{3}{c}{AVG mAP@IoU(\%)} \\
                \cline{4-6}
                 & && [0.1:0.5] & [0.3:0.7] & [0.1:0.7] \\
                \midrule
                CLIP-VIT B/32  &WIT-400M&-& 67.7 & 46.7 & 56.3 \\
                XCLIP-VIT B/32  &WIT-400M&Kinetics-400 & 68.5 & 48.1 & 57.4 \\
                CLIP-VIT B/16  &WIT-400M&-&68.9 & 48.3 & 57.7 \\
                XCLIP-VIT B/16  &WIT-400M&Kinetics-400  &69.6 & 49.3 & 58.5 \\
                \bottomrule
            \end{tabular}
        }
    }

    \label{caption feature extractor}
\end{table}%

\noindent\textbf{Impact of Caption Quality.}
In order to study the impact of caption quality on our method, we use different configurations of BLIP2 as the captioner to generate captions of different quality.
The results are present in Table~\ref{caption quality}. 
We observe that when BLIP2 utilizes the OPT with 6.7 B parameters as the language model, it achieves the best average mAP on various IoU thresholds.
It might be attributed to its sufficient amount of parameters, leading to higher-quality caption generation.
For this reason, we choose it as our caption generator.
\begin{table}[!tbp]
    \centering
    \caption{Performance with different configuration BLIP2.}
    \resizebox{\linewidth}{!}{
        \setlength{\tabcolsep}{1mm}{
            \begin{tabular}{l|c|ccccc}
                \toprule
                \multicolumn{1}{c|}{\multirow{2}{*}{Model}} & \multirow{2}{*}{\#Params} & \multicolumn{5}{c}{mAP@IoU(\%)} \\
                \cline{3-7}
                 & & 0.1 & 0.3 & 0.5 &0.7 & AVG\\
                \midrule
                BLIP2-OPT$_{2.7B}$ & 3.8B & 83.0 & 71.3&50.6&22.2 & 57.8 \\
                BLIP2-Flan-T5$_{xl}$ & 4.1B & 82.8 & 71.2&49.5&22.7 & 57.7 \\
                BLIP2-OPT$_{6.7B}$ & 7.8B & 83.6 & 71.6 &51.4&23.7& 58.5 \\
                \bottomrule
            \end{tabular}            
        }
    }
     \label{caption quality}
\end{table}%

\begin{figure}
\centering
\includegraphics[width=\linewidth]{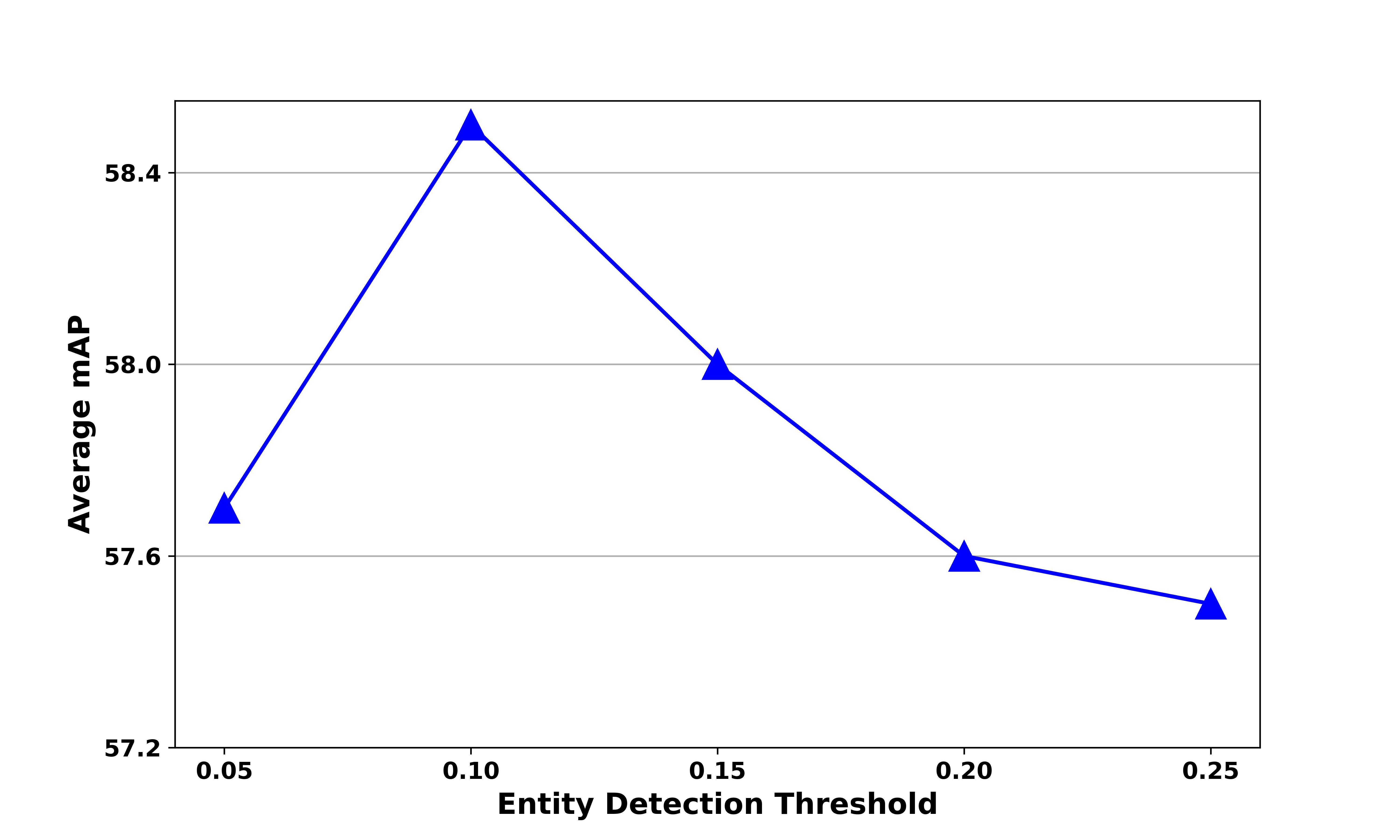}
\caption{Performance comparison (Average mAP for IoU thresholds of 0.1:0.7 ) under different entity detection thresholds.
}
\label{fig:entity_detection_threshold}

\end{figure}

\noindent\textbf{Influence of Different Entity Detection Threshold.}
The entity detection threshold $\gamma$ can partially affect detection performance as shown in Figure~\ref{fig:entity_detection_threshold}.
Both overly high and overly low entity detection thresholds can degrade the performance.
This is because a too-high threshold may ignore key entities, while a too-low threshold could introduce noisy entities (\textit{e.g.}, circular lights on a volleyball court).

\noindent\textbf{Ablation Study on Entity Refinement Strategies.}
\textcolor{black}{To investigate the individual contributions of the two entity refinement operations, we separately evaluate entity replacement and entity removal.
As shown in Table~\ref{tab:entity_refinement_ablation}, entity replacement alone achieves 69.2\%, 48.7\%, and 57.9\% on AVG mAP@[0.1:0.5], AVG mAP@[0.3:0.7], and AVG mAP@[0.1:0.7], respectively, while entity removal alone achieves 69.4\%, 48.2\%, and 57.7\%. Combining both operations yields the best results of 69.6\%, 49.3\%, and 58.5\%.
These results demonstrate that the two operations are complementary. Entity replacement corrects inaccurate entities, whereas entity removal filters out irrelevant ones. Their combination produces more reliable textual descriptions for subsequent multimodal alignment.}

\begin{table}[H]
\small
\centering
\caption{Ablation study on the entity refinement strategies.}
\label{tab:entity_refinement_ablation}
\resizebox{\linewidth}{!}{
\begin{tabular}{lccc}
\toprule
\textbf{Setting} &
\textbf{AVG mAP@[0.1:0.5]} &
\textbf{AVG mAP@[0.3:0.7]} &
\textbf{AVG mAP@[0.1:0.7]} \\
\midrule
Entity Replacement Only & 69.2 & 48.7 & 57.9 \\
Entity Removal Only     & 69.4 & 48.2 & 57.7 \\
Full Model              & {69.6} & {49.3} & {58.5} \\
\bottomrule
\end{tabular}
}
\end{table}

\noindent\textbf{Robustness to Noisy Captions.}
\textcolor{black}{To evaluate the robustness of TRA under noisy caption conditions, we conducted an additional experiment by intentionally corrupting the BLIP2-generated captions before applying PTR. Specifically, we randomly selected 10\% of the words in each caption and constructed two noisy settings by either masking or deleting the same selected words. As shown in Table~\ref{tab:caption_noise_robustness}, the average mAP@[0.1:0.7] only decreases from 58.5 with clean captions to 58.1 and 58.0 under word masking and deletion, respectively. 
These results show that noisy captions can affect localization performance, highlighting the importance of high-quality textual information. Nevertheless, TRA maintains relatively stable performance under different noise settings, demonstrating a certain degree of robustness.}

\begin{table}[H]
\small
\centering
\caption{Robustness analysis under noisy caption conditions.}
\label{tab:caption_noise_robustness}
\resizebox{\linewidth}{!}{
\begin{tabular}{lccc}
\toprule
\textbf{Setting} & \textbf{AVG mAP@[0.1:0.5]} & \textbf{AVG mAP@[0.3:0.7]} & \textbf{AVG mAP@[0.1:0.7]} \\
\midrule
10\% Masked & 69.3 & 49.1 & 58.1 \\
10\% Deleted & 69.1 & 48.8 & 58.0 \\
Clean & 69.6 & 49.3 & 58.5 \\
\bottomrule
\end{tabular}
}
\end{table}

\noindent\textbf{Effectiveness on Entity-Independent Actions.}
\textcolor{black}{
To evaluate the effectiveness of our method on entity-independent actions, we conducted experiments on the GTEA~\citep{gtea} dataset. This dataset mainly consists of action categories that do not depend on specific entities, such as ``take'' and ``open.'' These categories are primarily defined by the actions themselves rather than by particular interacting entities.
As shown in Table~\ref{tab:entity_independent_actions}, our method improves AVG mAP@[0.1:0.5], AVG mAP@[0.3:0.7], and AVG mAP@[0.1:0.7] by 8.2\%, 7.3\%, and 7.4\%, respectively. These results demonstrate that our method still achieves substantial performance gains when entity-specific cues are limited. This is mainly because TRA introduces textual semantic features generated from video content and enhances action representations through visual-language alignment. For such entity-independent actions, even when little entity replacement or deletion is required, the original captions can still provide effective action-level semantic information to facilitate action localization. Therefore, TRA is also effective for entity-independent actions.
}

\begin{table}[H]
\centering
\caption{Performance on the entity-independent GTEA dataset.}
\label{tab:entity_independent_actions}
\resizebox{0.8\linewidth}{!}{
\begin{tabular}{lcc}
\toprule
\textbf{Metric} & \textbf{Baseline} & \textbf{Ours} \\
\midrule
AVG mAP@[0.1:0.5] & 55.1 & 63.3 \\
AVG mAP@[0.3:0.7] & 36.8 & 44.1 \\
AVG mAP@[0.1:0.7] & 45.4 & 52.8 \\
\bottomrule
\end{tabular}
}
\end{table}

\subsection{Discussion on Computational Cost}
\begin{table}[ht]
\centering
\caption{Computational cost analysis of each pre-trained model during inference (including GPU memory usage from model parameters and forward propagation), measured in MB on one NVIDIA GeForce RTX-3090.}
\resizebox{0.5\textwidth}{!}{%
\begin{tabular}{ccccccccccc}
\toprule
Model & BLIP2 & TextGraphParser & OWLv2 & AngLE & XCLIP  \\
\midrule
GPU Memory Usage (MB)
 & 16656 & 2158 & 8492 & 2374 & 2760 \\
\bottomrule
\end{tabular}%
}
\label{tab:computational cost}
\end{table}

\begin{table}[H]
\centering
\caption{Runtime comparison between the baseline and TRA.}
\label{tab:runtime_comparison}
\resizebox{0.5\textwidth}{!}{
\begin{tabular}{lccc}
\toprule
\textbf{Method} & \textbf{Training  (min)}
& \textbf{Inference  (min)}
& \textbf{PTR  (min/1-min video)} \\
\midrule
Baseline & 15.67 & 0.50 & -- \\
TRA      & 16.89 & 0.53 & 19.81 \\
\bottomrule
\end{tabular}
}
\end{table}

Our framework involves several pre-trained models for textual semantic enhancement compared to the baseline, resulting in a certain increase in computational cost.
Table~\ref{tab:computational cost} presents the GPU memory usage of each pre-trained model during inference, ranging from 2158MB to 16656MB.
However, these pre-trained models are invoked sequentially rather than concurrently. As a result, the entire pipeline can be executed on a single GPU, such as a consumer-grade NVIDIA GeForce RTX-3090 with 24GB of memory, making the overall computational cost manageable and acceptable.

\textcolor{black}{
We also compare the runtime of TRA with that of the baseline under the same experimental setting. As shown in Table~\ref{tab:runtime_comparison}, when considering only the training and inference processes of the temporal action localization model, the training time increases from 15.67 min to 16.89 min, while the inference time increases only slightly from 0.50 min to 0.53 min. This indicates that TRA introduces only limited additional time overhead during training and inference. In addition, PTR requires 19.81 min to preprocess a one-minute video, resulting in additional preprocessing time. 
}

\begin{figure*}[!t]
\centering
\includegraphics[width=\linewidth]{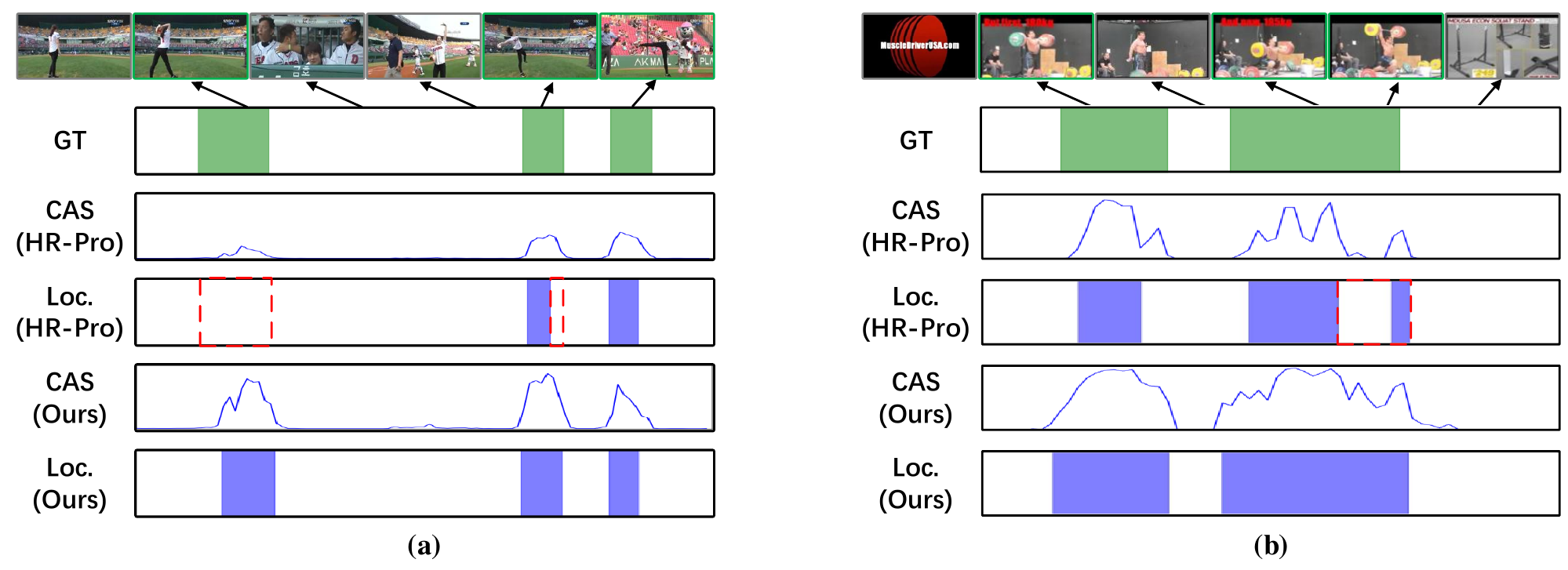}

\caption{
\textcolor{black}{
Qualitative comparison between our proposed method and HR-Pro~\citep{HR-Pro} on THUMOS'14. We present two examples from different action classes: (a) BaseballPitch and (b) CleanAndJerk. The first row shows the input video, followed by the ground-truth intervals. The subsequent rows present the class-specific activation sequences (CASs) and localization results of HR-Pro and our method. Red boxes indicate false positives or missed detections. In (a), our method detects more complete action instances with clearer boundaries, while in (b), it avoids the fragmented prediction produced by HR-Pro. Overall, our results achieve better temporal alignment with the ground truths.
}
}
\label{fig:quality_result1}

\end{figure*}


\subsection{Qualitative Results}

\noindent\textbf{Qualitative Comparison.}
In Figure~\ref{fig:quality_result1}, we compare the qualitative result of our method with HR-Pro on test videos in THUMOS’14.
Figure~\ref{fig:quality_result1}(a) shows that our model provides more precise localization for action instances.
It is evident that HR-Pro ignores the first action instance, while the second instance is predicted incompletely, with some parts missing.
By contrast, our method detects more complete snippets and achieves clearer boundaries than HR-Pro.
In Figure~\ref{fig:quality_result1}(b), HR-Pro incorrectly divides the second action instance into two parts, whereas our method successfully detects the complete instance.


\subsection{Limitations}
\begin{figure}
\centering
\includegraphics[width=\linewidth]{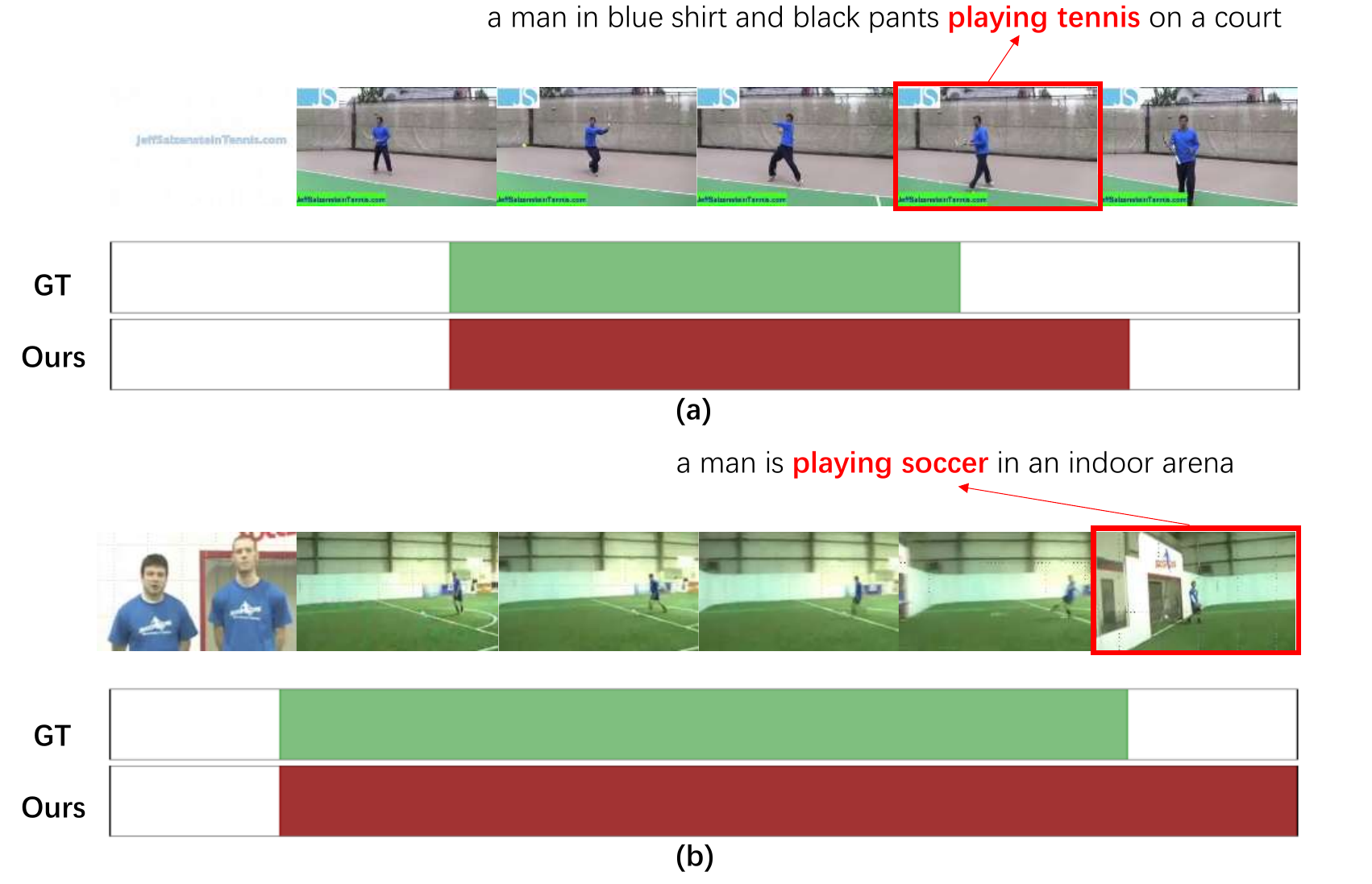}

\caption{Visualization of representative failure cases for (a) \textit{TennisSwing} and (b) \textit{SoccerPenalty}. The red boxes highlight false-positive predictions, where background segments near the action boundaries are incorrectly classified as parts of the corresponding action instances.
}
\label{fig:failure_cases}

\end{figure}

\textcolor{black}{Although TRA effectively refines generic captions and aligns the visual and linguistic modalities, thereby enriching semantic information and improving localization precision, it remains limited in resolving fine-grained temporal boundary ambiguity. As illustrated in Figure~\ref{fig:failure_cases}(a), the model correctly localizes the main \textit{TennisSwing} process but mistakenly includes the motion after the swing. The refined caption still describes this segment as ``a man in a blue shirt and black pants playing tennis on a court,'' which remains highly relevant to the target action. Similarly, in Figure~\ref{fig:failure_cases}(b), the segment following \textit{SoccerPenalty} is incorrectly identified as part of the action, while the generated caption, ``a man is playing soccer in an indoor arena,'' remains consistent with the overall action context. Since these captions contain no explicit incorrect or irrelevant entities, the proposed entity replacement and deletion operations cannot effectively eliminate such boundary confusion. These cases indicate that the current refinement strategy mainly addresses entity-level semantic errors but remains limited when action and neighboring segments exhibit highly similar visual and contextual information.
To address this limitation, future work will explore finer-grained text refinement beyond entity-level correction by modeling action phases and subtle motion changes, thereby improving the distinction between target actions and similar neighboring segments.
}


\textcolor{black}{
Beyond the aforementioned limitation, TRA faces two practical challenges. First, its text refinement pipeline relies on multiple pretrained models for caption generation, parsing, and refinement, resulting in additional preprocessing costs that may hinder its application to large-scale or real-time scenarios. A promising solution is to use the refined captions to adapt the original captioner via parameter-efficient fine-tuning methods, such as LoRA. This could transfer the refinement capability to a lightweight captioner and remove the need for multiple pretrained models during deployment. Second, the action-entity mappings produced by ChatGPT-4 may contain hallucinated or ambiguous entities and thus require manual correction. As this correction is performed at the category level, its cost grows with the number of action categories, potentially limiting scalability on datasets with numerous fine-grained classes. Future work will investigate more reliable and automated LLM-based mapping and confidence-aware verification strategies to minimize or eliminate manual intervention.
}

\section{Conclusion}
This work proposes a novel \textcolor{black}{Text Refinement and Alignment (}TRA\textcolor{black}{)} framework for point-supervised temporal action localization, which \textcolor{black}{incorporates} refined captions and aligns visual and linguistic modalities to \textcolor{black}{provide complementary semantic information for more accurate action localization}.
The proposed PTR \textcolor{black}{uses the descriptions associated with point-annotated frames as references} and leverages several pre-trained models to \textcolor{black}{refine inaccurate entities in the initial captions}.
Subsequently, the PMA uses the \textcolor{black}{pseudo semantic labels derived from the predicted action and background points} to align \textcolor{black}{visual and textual representations}.
Extensive experiments on five benchmarks demonstrate that TRA \textcolor{black}{consistently improves} existing PTAL methods and achieves state-of-the-art \textcolor{black}{performance, validating the effectiveness and generalizability of the proposed framework}.
\textcolor{black}{Beyond the immediate performance improvements, these results indicate that refined textual descriptions can provide effective complementary semantic information under sparse point supervision, offering a promising direction for reducing the reliance on dense temporal annotations.}
\textcolor{black}{In the long term, this work may promote the development of multimodal learning paradigms for weakly supervised video understanding and facilitate the use of textual semantic guidance in other related video analysis tasks.}
\textcolor{black}{Future work will focus on reducing the text preprocessing cost and improving fine-grained temporal modeling to enhance the scalability and practical applicability of the framework.}

\bibliographystyle{model5-names}

\bibliography{cas-refs}

\end{document}